\documentclass[10pt,twocolumn,letterpaper]{article}

\usepackage{cvpr}              % To produce the CAMERA-READY version
\usepackage{graphicx}
\usepackage{amsmath}
\usepackage{amssymb}
\usepackage{booktabs}
\usepackage{soul}
\usepackage{enumitem}
\usepackage{multirow}                   
\usepackage[accsupp]{axessibility}  % Improves PDF readability for those with disabilities.

\usepackage[pagebackref,breaklinks,colorlinks]{hyperref}

\usepackage[capitalize]{cleveref}
\crefname{section}{Sec.}{Secs.}
\Crefname{section}{Section}{Sections}
\Crefname{table}{Table}{Tables}
\crefname{table}{Tab.}{Tabs.}

\def\cvprPaperID{9788} % *** Enter the CVPR Paper ID here
\def\confName{CVPR}
\def\confYear{2023}

\begin{document}

\title{\textit{Fantastic Breaks:} A Dataset of Paired 3D Scans of Real-World Broken Objects and Their Complete Counterparts}

\author{    
Nikolas Lamb, Cameron Palmer, Benjamin Molloy, Sean Banerjee, Natasha Kholgade Banerjee\\
Clarkson University, Potsdam NY, USA\\
{\tt\small \{lambne, campalme, molloybr, sbanerje, nbanerje\}@clarkson.edu}
% First Author\\
}

\maketitle

%%%%%%%%% ABSTRACT
\begin{abstract}
  Automated shape repair approaches currently lack access to datasets that describe real-world damaged geometry. We present \emph{Fantastic Breaks (and Where to Find Them: \url{https://terascale-all-sensing-research-studio.github.io/FantasticBreaks})}, a dataset containing  scanned, waterproofed, and cleaned 3D meshes for 150 broken objects, paired and geometrically aligned with complete counterparts. \emph{Fantastic Breaks} contains class and material labels, proxy repair parts that join to broken meshes to generate complete meshes, and manually annotated fracture boundaries. Through a detailed analysis of fracture geometry, we reveal differences between \emph{Fantastic Breaks} and synthetic fracture datasets generated using geometric and physics-based methods. We show experimental shape repair evaluation with \emph{Fantastic Breaks} using multiple learning-based approaches pre-trained with synthetic datasets and re-trained with subset of \emph{Fantastic Breaks}.
\end{abstract}

%%%%%%%%% BODY TEXT

\section{Introduction}
\label{sec:intro}

Damage to objects is an expected occurrence of everyday real-world usage. However, when damage occurs, objects that could be repaired are often thrown out. Additive manufacturing techniques are rapidly becoming accessible at the consumer level, with 3D printing technologies available for materials such as plastics, metals, and even ceramics and wood. Though current approaches for repair have been largely manual and restricted to niche areas such as cultural heritage restoration, a large body of recent research has emerged on the automated reversal of damage, including reassembly of fractured parts using 3D scans~\cite{brown2008system,funkhouser2011learning,papaioannou2001virtual,papaioannou2003automatic,huang2006reassembling,zhang20153d,hong2019potsac,hong2021structure,zhan2020generative,wu2020pq,harish2022rgl,chen2022neural}, or generation of new repair parts when portions of the original object are irretrievably lost due to damage~\cite{li2011symmetry, sipiran2014approximate, gregor2015automatic, mavridis2015object, papaioannou2017reassembly, sipiran2018completion, hermoza20183d, lamb2022mendnet, lamb2022deepmend, lamb2022deepjoin}. Geometry-driven approaches based on shape matching~\cite{li2011symmetry, sipiran2014approximate, gregor2015automatic, mavridis2015object, papaioannou2017reassembly, sipiran2018completion, lamb2019automated, brown2008system, funkhouser2011learning, papaioannou2001virtual, papaioannou2003automatic, huang2006reassembling, hong2019potsac, hong2021structure} that are not usable for objects of unknown complete geometry have given way to learning-driven approaches~\cite{zhan2020generative, wu2020pq, harish2022rgl, chen2022neural, hermoza20183d, lamb2022mendnet, lamb2022deepmend, lamb2022deepjoin} aimed at generalization to repair at a large scale. 

\begin{figure}
    \centering
    \includegraphics[width=\linewidth]{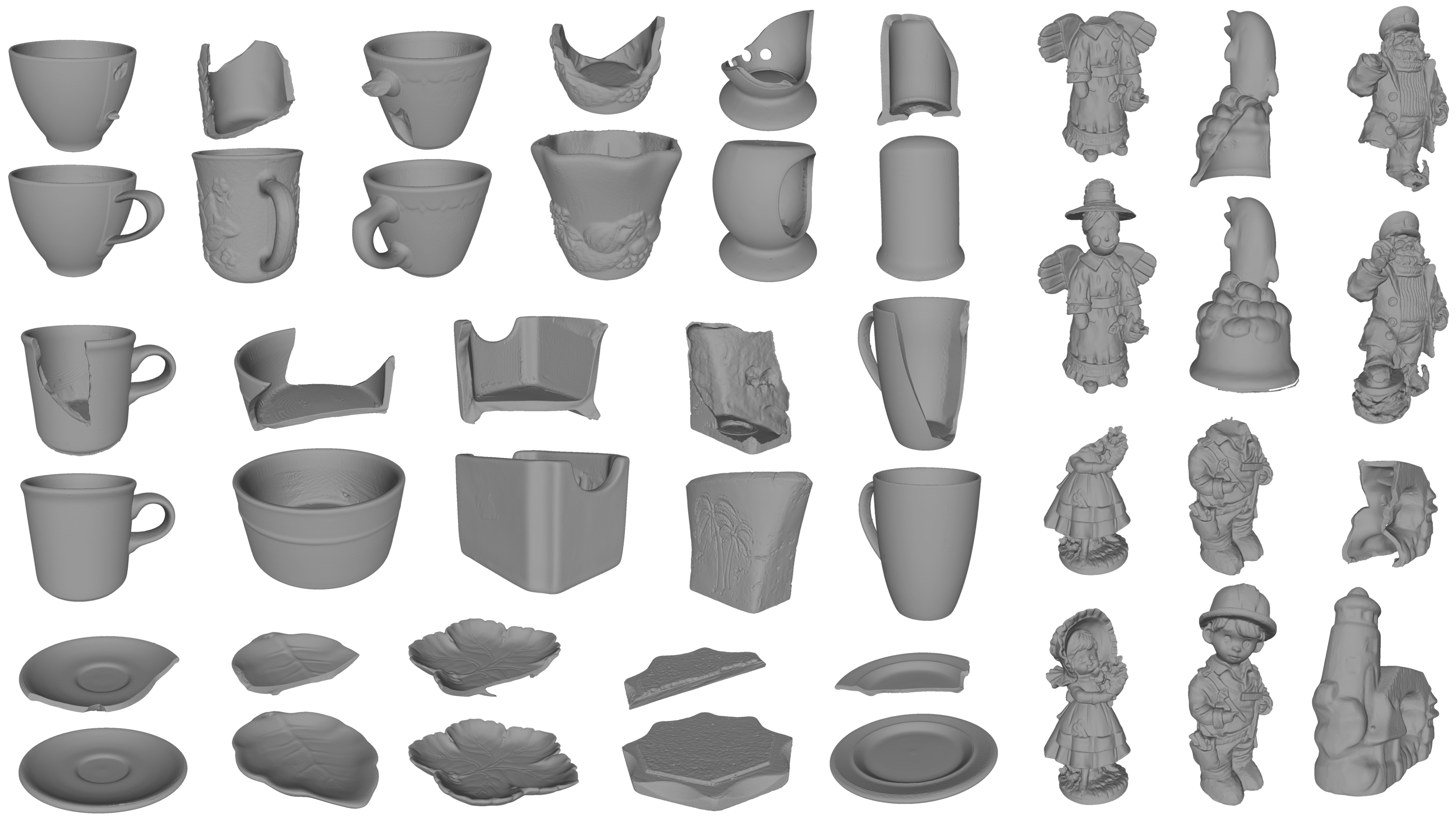}
    \caption{We present \textit{Fantastic Breaks}, a dataset of 3D scans of real-world broken objects (top) aligned to 3D scans of complete counterparts (bottom). Objects span classes such as mugs, plates, statues, jars, and bowls\textemdash{}household items prone to damage.}
    \label{fig:teaser}
\end{figure}

However, a principal challenge limiting understanding of real-world damage and limiting application-focused evaluation of repair approaches is that datasets of real-world damage for consumer space objects are virtually non-existent. Current learning-driven approaches for repair use datasets where fracture-based damage is synthetically generated using geometric approaches such as Boolean operations with primitives~\cite{gregor2015automatic, lamb2021using, lamb2022mendnet, lamb2022deepmend, lamb2022deepjoin, chen2022neural}. As they make assumptions about the fracture process rather than using data-driven fracture generation, the practical usability of such geometric methods for repairing real fractures is unknown. With Breaking Bad, Sell\'{a}n et al.~\cite{sellan2022breaking} have taken a first step toward large-scale fracture dataset generation. Breaking Bad consists of 3D shapes from Thingi10k~\cite{zhou2016thingi10k} and PartNet~\cite{mo2019partnet} subjected to physics-based damage using fracture modes~\cite{sellan2022breakinggood}. By removing macro-scale shape assumptions embodied by geometric primitives, Breaking Bad is a promising step for research in shape assembly and repair. However, fractures in Breaking Bad suffer from typical issues of resolution and simulation time step size that underlie physics simulations. The dataset is thus unfortunately hindered in providing a faithful representation of real-world damage. 

In this work, we contribute \textit{Fantastic Breaks}, the first dataset of 3D scans of damaged objects paired with 3D scans of their complete non-damaged counterparts, as shown in Figure~\ref{fig:teaser}. Each damaged object\textemdash{}hereafter referred to a broken object due to the nature of damage suffered\textemdash{}is 3D scanned and geometrically registered to a scan of the undamaged complete object, such that intact regions of the complete and broken scans are aligned. At the time of publication, the dataset contains 150 broken/complete pairs. 

Our damage infliction often leaves one broken part intact, destroying or over-fragmenting the remainder of the object. We use an off-the-shelf subtraction-based approach~\cite{lamb2019automated} to generate repair part proxies from the intact broken part aligned to its complete counterpart. We also provide manually annotated class labels, material labels, and annotated fractured regions on the broken meshes. Our work emulates endeavors of groups in vision and robotics that contribute datasets of 3D scanned everyday-use objects~\cite{singh2014bigbird,calli2015benchmarking,calli2017yale,kasper2012kit,downs2022google}. Our analysis of real-world fracture properties in \textit{Fantastic Breaks} reveals fine-scale fracture structure that enables our dataset to overcome the drawbacks of synthetic datasets. We use our dataset to evaluate existing shape repair approaches on real fractured objects.

We summarize our contributions as follows:

\begin{enumerate}[noitemsep,topsep=0pt]
    \item We contribute the first 3D scanned real-world dataset of geometrically aligned broken/complete object pairs to enable application-focused evaluation of repair.
    \item We provide class, material, and fracture surface annotations, and ground truth repair part proxies.
    \item We contribute a geometric analysis of \textit{Fantastic Breaks} in comparison to existing synthetic fracture datasets.
    \item We provide evaluations of existing shape repair approaches using \textit{Fantastic Breaks}.
\end{enumerate}

\section{Related Work}
\label{sec:relatedwork}

\paragraph{Fracture Datasets.} Existing real-world fracture collections are restricted to scanning of multiple shards corresponding to a small set of originally intact objects, e.g., 7 objects~\cite{huang2006reassembling}, 3 frescoes~\cite{brown2008system}, and 3 large-scale structures (Akrotiri settlement, Tongeren Roman excavation, and one fresco)~\cite{funkhouser2011learning}. Since fracture acquisition is performed post-damage for historical objects where no known counterparts exist, the datasets lack knowledge of the complete proxies. The Hampson Museum cultural heritage dataset~\cite{payne2009designing} contains 3D scans for 138 cultural heritage objects. The dataset lacks paired damaged/complete data, or annotations to reveal what objects are damaged and what are intact, preventing them from being used to train repair or assembly approaches. Hong et al.~\cite{hong2021structure} show assembly results using a dataset of 5 shattered pots, with 3D scans acquired for the shattered fragments and the original pots. The original scans are used as evaluation oracles. Lamb et al.~\cite{lamb2019automated} repair 22 damaged objects by subtracting the damaged objects from \textit{a priori} known complete proxies. 

Recognizing the need for large-scale datasets for learning-driven repair, a few datasets contain synthetic or scanned models subjected to synthetic fracture using geometric techniques such as subtracting primitives~\cite{gregor2015automatic,lamb2021using,lamb2022mendnet,lamb2022deepmend,lamb2022deepjoin,chen2022neural}, or using physics models of fracture~\cite{sellan2022breaking}. As we demonstrate in Section~\ref{sec:analysis}, real-world physical damage demonstrates geometric characteristics that differ from the break patterns of synthetically generated damage. Geometric fracture models~\cite{gregor2015automatic,lamb2021using,lamb2022mendnet,lamb2022deepmend,lamb2022deepjoin,chen2022neural} are only as precise as the primitive being used for fracture, e.g., Chen et al.~\cite{chen2022neural} use five simplistic cut functions\textemdash{}planar, sine, parabolic, square, and pulse\textemdash{}modeled as oriented height fields. Lamb et al.~\cite{lamb2021using,lamb2022mendnet,lamb2022deepmend,lamb2022deepjoin} subtract randomly rotated and translated geometric primitives such as a cube, icosphere, and sub-divided icosphere, with and without random surface perturbation to simulate micro-scale detail. Gregor et al.~\cite{gregor2015automatic} use spheres whose surfaces are perturbed using small-scale details of a single digitized material. In all cases, mid-scale detail is represented as cutouts via analytical primitives, which is not generalizable to real-world damage. 

Though approaches exist to perform fracturing of single objects using physics simulations~\cite{fan2022simulating, chitalu2020displacement, joshuah2020anisompm, wolper2019cd, mandal2022remeshing}, which could theoretically be used to generate a fractured object dataset, they suffer from drawbacks that prevent them from being used at scale.
\textit{(1)~Unrepairable Objects.} Objects are repairable only if a large part remains intact. Approaches that cannot control the number of parts~\cite{fan2022simulating, chitalu2020displacement, joshuah2020anisompm, wolper2019cd} may require many trials to obtain a repairable object.
\textit{(2)~Numerical Instability.} Polygon division may cause simulation failure due to roundoff error~\cite{chitalu2020displacement, mandal2022remeshing}.
\textit{(3)~Limited Diversity.} Some approaches tend to produce planar fractures~\cite{fan2022simulating} or limit crack speed to ensure PDE convergence~\cite{joshuah2020anisompm}.
\textit{(4)~Runtimes.} Fractures on objects with similar degrees of freedom to ours (2.5M) require multiple hours, e.g. 6h/1M~\cite{fan2022simulating}, 7h/2.5M~\cite{joshuah2020anisompm}, 3.5h/2.5M~\cite{mandal2022remeshing}, 26h/2.5M~\cite{chitalu2020displacement}, 4h/1M~\cite{wolper2019cd}.
\textit{(5)~Parameter Tuning.} Physics parameters, e.g. density, fracture threshold, and timestep size~\cite{joshuah2020anisompm}, need tuning to produce diverse fractures, a process which has a ``notable learning curve''~\cite{wolper2019cd}. In summary, excessive parameter tuning and long runtimes make dataset creation using physics simulations intractable. The only physics-based fractured dataset is Breaking Bad~\cite{sellan2022breaking}. The Breaking Bad dataset of Sell\'{a}n et al. uses their earlier work~\cite{sellan2022breakinggood} to model and simulate fracture modes of the object. Their approach struggles to generate high-resolution detail, as it requires evolving the simulation over small time steps that can be computationally infeasible at a large scale. Our \textit{Fantastic Breaks} dataset fills the gap in the lack of real-world damaged object datasets by contributing an even-now growing repository of 3D scanned real-world damaged objects rigidly aligned with 3D scans of their real-world counterparts.

\paragraph{Real-World Object Datasets.} Capture of large-scale complete 3D scans of objects is an arduous task, due to the need for a multi-staged approach consisting of multiple presentations of the object to a scanner, registration of scans, cleaning to correct imprecise geometry, eliminate holes, and correct deep concavities, and, depending on the application, waterproofing to ensure closed surfaces. Massive 3D datasets typically comprise single-viewpoint RGB-D images~\cite{janoch2013category,lai2013rgb,xiang2016objectnet3d}, room-scale 3D scans~\cite{dai2017scannet}, or scans of large objects such as motorcycles, large statues, and benches~\cite{choi2016large} that cannot be readily impacted to capture real-world damage. The Berkeley BigBIRD dataset~\cite{singh2014bigbird} contained 100 3D models at publication time and has grown to 125 models. Objects were recorded on a turntable and models were created by stitching multiple Kinect and Canon camera images. Using the BigBIRD setup, Calli et al.~\cite{calli2015benchmarking,calli2017yale} collected the YCB dataset for a set of 77 objects as a collaboration between Yale, CMU, and UC Berkeley. Due to placement on a turntable, hidden portions, e.g., the base or concavities not visible to the cameras, are not captured. The Karlsruhe Institute of Technology (KIT) dataset~\cite{kasper2012kit} consists of 145 objects captured using a Konica Minolta Vi-900 digitizer, with multiple scans acquired to capture object bases. With 300 3D scans (150 broken, 150 paired complete counterparts, and growing), \textit{Fantastic Breaks} is comparable to these prior tabletop datasets~\cite{singh2014bigbird,calli2015benchmarking,calli2017yale,kasper2012kit} in terms of total items.

Large-scale table-top 3D scanned object datasets are few in number. Google Scanned Objects (GSO)~\cite{downs2022google} consists of 1,030 3D scans of household objects. GSO is collected by 3D imaging projector-cast patterns through a collaboration between Google and Open Robotics. Another large-scale dataset, AKB-48~\cite{liu2022akb}, contains 2,037 3D scans of complete objects. Scans are obtained using an Einscan 3D scanner and an Intel RealSense depth camera. Unlike \textit{Fantastic Breaks}, neither GSO nor AKB-48 contain broken objects.

Within the larger problem domain of minimizing waste, a few 2D datasets have arisen for object detection and segmentation in waste images~\cite{sousa2019automation,proencca2020taco,bashkirova2022zerowaste}. Object materials span cardboard, plastic, glass, and metal. An opportunity exists to capture RGB-D images of waste, identify objects capable of being repaired, and geometrically couple them with our dataset to conduct in-the-wild repair.

\paragraph{Shape Repair.} The creation of the \textit{Fantastic Breaks} dataset is motivated by applications in object repair. When object parts are available, shape assembly approaches focus on joining 3D shape representations of the object parts. Geometric approaches exist to match fracture boundaries via segmentation~\cite{papaioannou2001virtual,papaioannou2003automatic}, feature description extraction and geometric model refinement~\cite{huang2006reassembling}, align fractured shapes to a proxy template~\cite{zhang20153d}, or conduct iterative registration similar to structure from motion~\cite{hong2019potsac,hong2021structure}. Several learning-based approaches exist to provide assembly, most of which assume holistic parts with simple surfaces~\cite{zhan2020generative,wu2020pq,harish2022rgl}, though one assumes arbitrary geometry at shape boundaries~\cite{chen2022neural}.

When parts of an object are irretrievable, shape repair approaches address generation of the lost parts. Early automation approaches to circumvent historically manual repair included finding symmetries and self-similarities in objects~\cite{li2011symmetry,sipiran2014approximate,gregor2015automatic,mavridis2015object,papaioannou2017reassembly,sipiran2018completion}, though these approaches are unsuccessful when non-symmetric object parts are broken off. One approach~\cite{lamb2019automated} circumvents the small-scale artifact issue of Boolean subtraction by automatically extracting and joining exterior and fractured regions for repair parts using real-world damaged and complete scans. The approach requires the scan of a complete 3D proxy to be provided as input, that may not be feasible for one-of-a-kind instances, or even if available, may prove tedious to obtain. 

Recent work such as 3D-ORGAN~\cite{hermoza20183d}, MendNet~\cite{lamb2022mendnet}, DeepMend~\cite{lamb2022deepmend}, and DeepJoin~\cite{lamb2022deepjoin} uses deep learning to conduct shape repair without knowledge of the complete proxy by representing damaged, complete, and repair shapes using voxels~\cite{hermoza20183d} or deep functions~\cite{lamb2022mendnet,lamb2022deepmend,lamb2022deepjoin}. DeepMend and DeepJoin report higher success due to the use of implicit functions which enable representation of arbitrary resolutions, and due to the expression of fractured and restoration shapes in terms of constructive solid geometry operations between the complete object and a break surface whose representation is learnt during training. Though DeepMend and DeepJoin show qualitative results on a few examples of 3D scans of real-world damaged objects, in all cases, training and quantitative evaluation is conducted using datasets with synthetically-generated fractures.

\section{Data Collection and Processing}
\label{sec:datacoll}

\begin{figure}[t!]
    \centering
    \includegraphics[width=\linewidth]{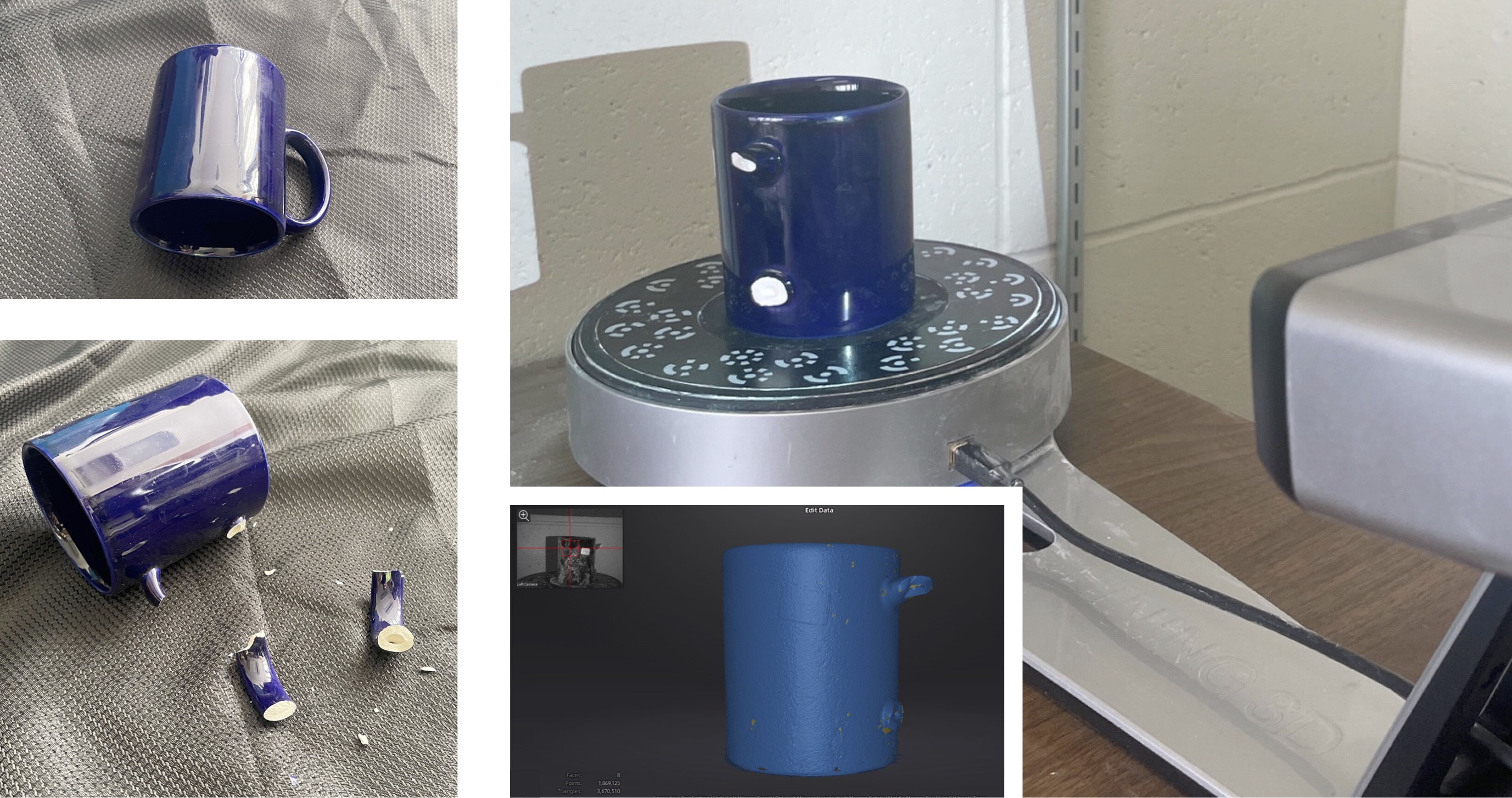}
    \caption{Left: Complete mug on top, and broken mug on the bottom showing main intact part and the shattered handle. Right: Object on scanner with 3D scan shown in the inset.}
    \label{fig:fracturescan}
\end{figure}

\begin{figure}[t!]
    \centering
    \includegraphics[width=\linewidth]{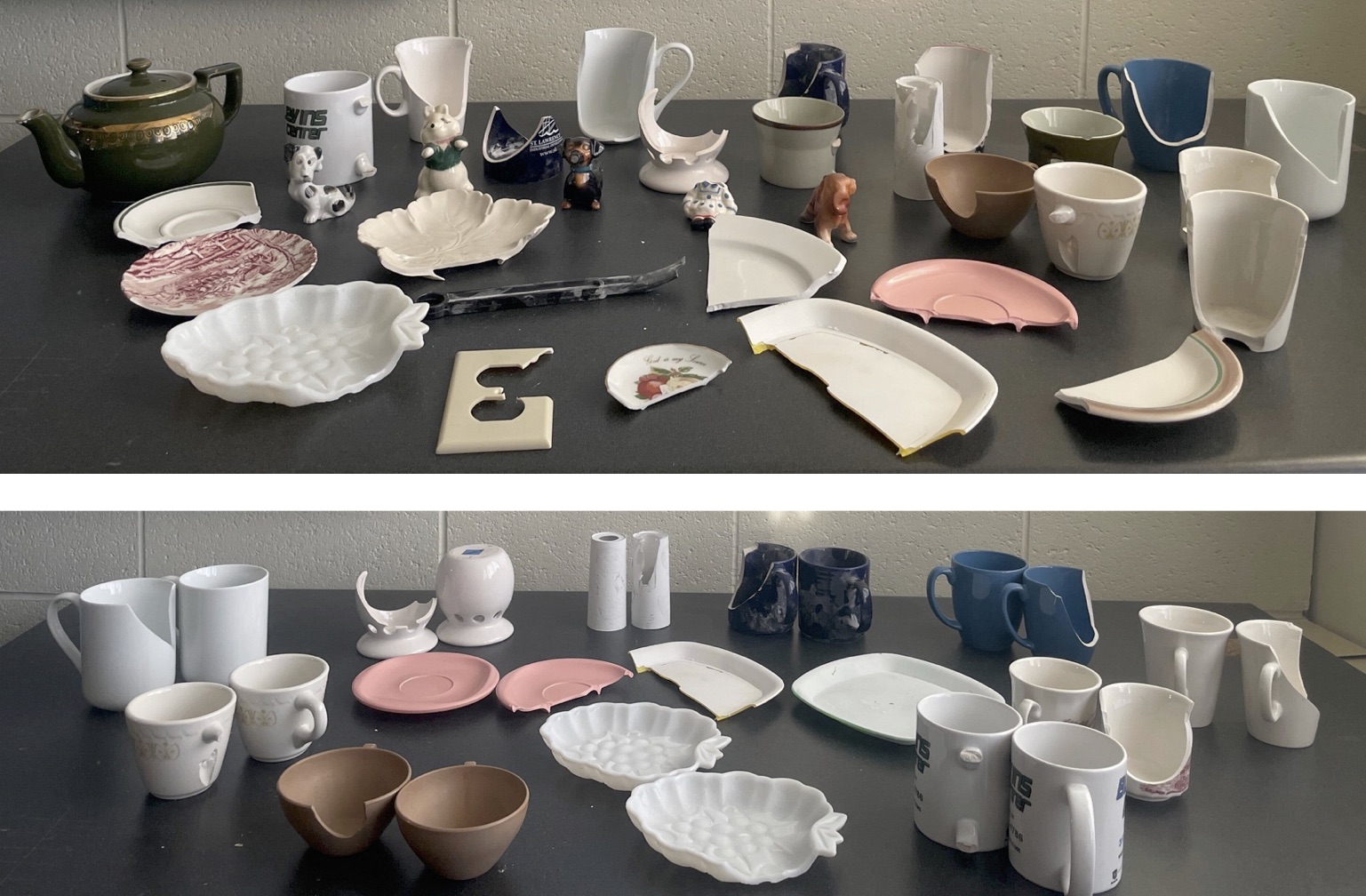}
    \caption{Example broken objects on top and broken/complete pairs on the bottom acquired to build the \textit{Fantastic Breaks} dataset.}
    \label{fig:collection}
\end{figure}

\textbf{Object Acquisition.} We conducted a community-wide acquisition of everyday household objects that suffer damage. Our goal was to have a collection consisting not only of objects that can be damaged, but also of objects that have already suffered damage. To perform our acquisition, we made purchases at the local thrift store where we found intact and damaged objects, and requested donations from the local community. Though the \textit{Fantastic Breaks} dataset as presented in this paper contains damaged/complete pairs, we have also collected pre-damaged objects that may lack complete counterparts, as they provide insight into real-world fracture. Our collection consists of commonly damaged household objects such as mugs, plates, and statues spanning materials such as ceramics, plastics, glass, and wood. In some instances, we were able to pre-acquire pairs, where one object in the pair was damaged and the other was intact. In most cases, we manually damaged a complete object to obtain the broken version. When possible, we attempted to acquire pairs of complete objects, and damage one of the objects, enabling us to store physical damaged/complete pairs. We inflicted damage by dropping the object, striking the object with a rubber mallet or metal hammer, and, in 3 cases, snapping the object after anchoring it against a table. Figure~\ref{fig:fracturescan} shows an example mug with the handle shattered after striking with a mallet. Figure~\ref{fig:collection} shows broken objects and broken/complete pairs in our dataset.

\textbf{3D Scanning.} We use an Einscan SP 3D turntable-based scanner to acquire 3D scans. We use Einscan's proprietary software, EXScan, to operate the scanner. The scanning operation rotates the scanner's turntable 8 times, acquires 8 2.5D images of the object via an attached RGB-D sensor, and fuses the images into a 3D mesh. Given the diversity of object geometry where objects may contain deep concavities or complex fractures, we had to conduct careful staging of each object and perform multiple presentations of the object to maximize acquisition of the object surface. The presentation count ranged from 2 for flat objects such as plates or objects with higher convexity such as statues, to 6 for objects with concavities such as mugs. We used the registration tool in EXScan to fuse scans into a 3D mesh.

\begin{figure}[t!]
    \centering
    \includegraphics[width=\linewidth]{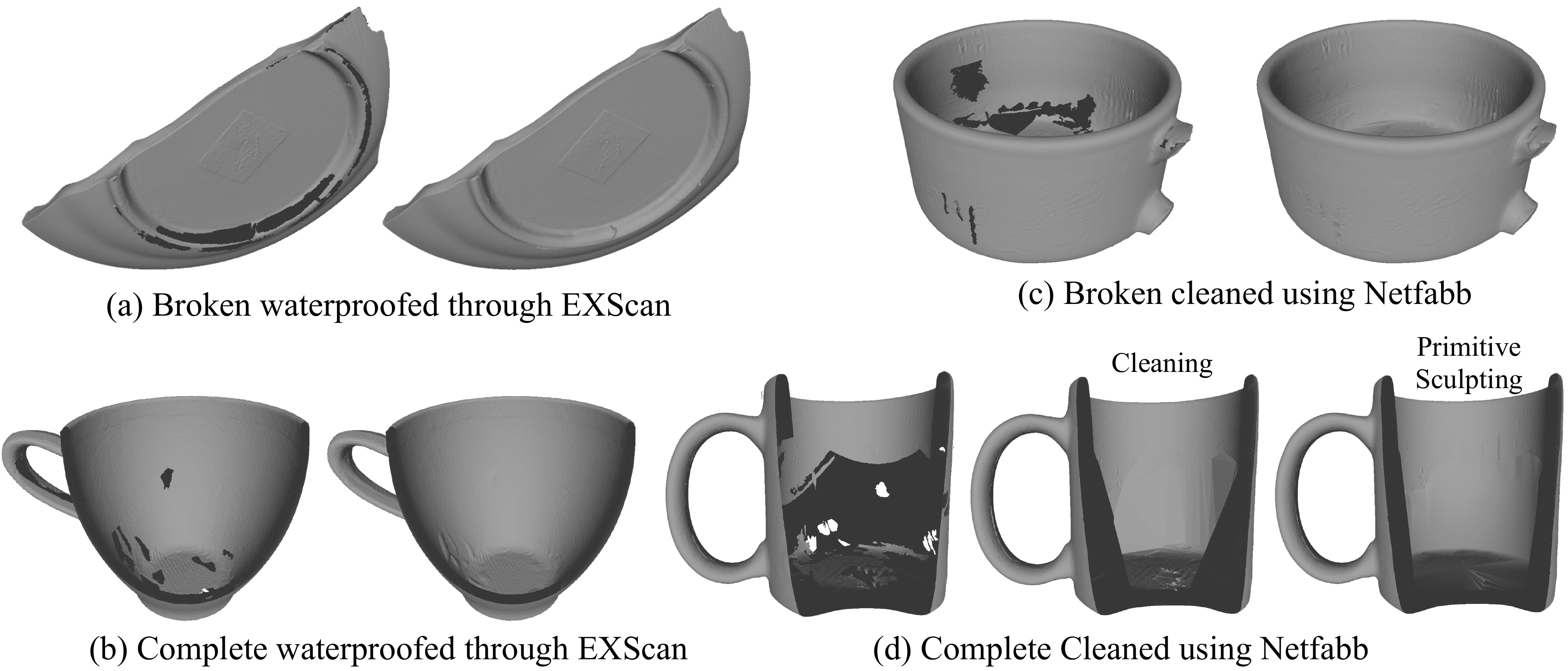}
    \caption{Meshes prior to (left) and after cleaning and waterproofing for (a)~broken and (b)~complete object examples waterproofed using EXScan, and (c)~broken and (d)~complete objects cleaned using Netfabb. For (d), we use a primitive to fix the erroneously angled surfaces created by Netfabb mesh repair tools. Complete objects are shown as cutaways to reveal pre-cleaning artifacts.}
    \label{fig:cleaning}
\end{figure}

\textbf{Mesh Cleaning.} Given the 3D scan of a complete or broken object, we employ a sequence of operations to ensure that the meshes are of high quality. We visually inspect each model, and if the model shows any small holes in the mesh, we use the waterproofing tool built into EXScan to ensure that each mesh is a closed 2D surface. While performing waterproofing in EXScan is preferred, if the model contains large holes or artifacts, we repair the mesh by performing manual hole-filling using Autodesk Netfabb. For objects with deep concavities such as cups and mugs, the concavities may not have been well-presented during the scanning process, in which case the interior regions may not be metrically accurate. If possible, we repair the mesh using subtraction with a geometric primitive, e.g., a cylinder. We apply the ``Extended Repair'' scripts in Netfabb to merge nearby vertices, remove double and flipped triangles, close all holes, wrap the mesh to remove interior faces, and remove small connected components. Figure~\ref{fig:cleaning} demonstrates examples of cleaned meshes.

\textbf{Mesh Orientation.} After cleaning, we manually orient each mesh such that its principal axes are aligned to the Cartesian axes. We align the base of the object with the \textit{xz}-plane, and we rotate the object about the \textit{y}-axis to ensure alignment within its category. For example, we align mugs such that the handles are aligned with the negative \textit{x} direction, and statues so that they face in the negative \textit{x} direction.

\begin{figure}[t!]
    \centering
    \includegraphics[width=\linewidth]{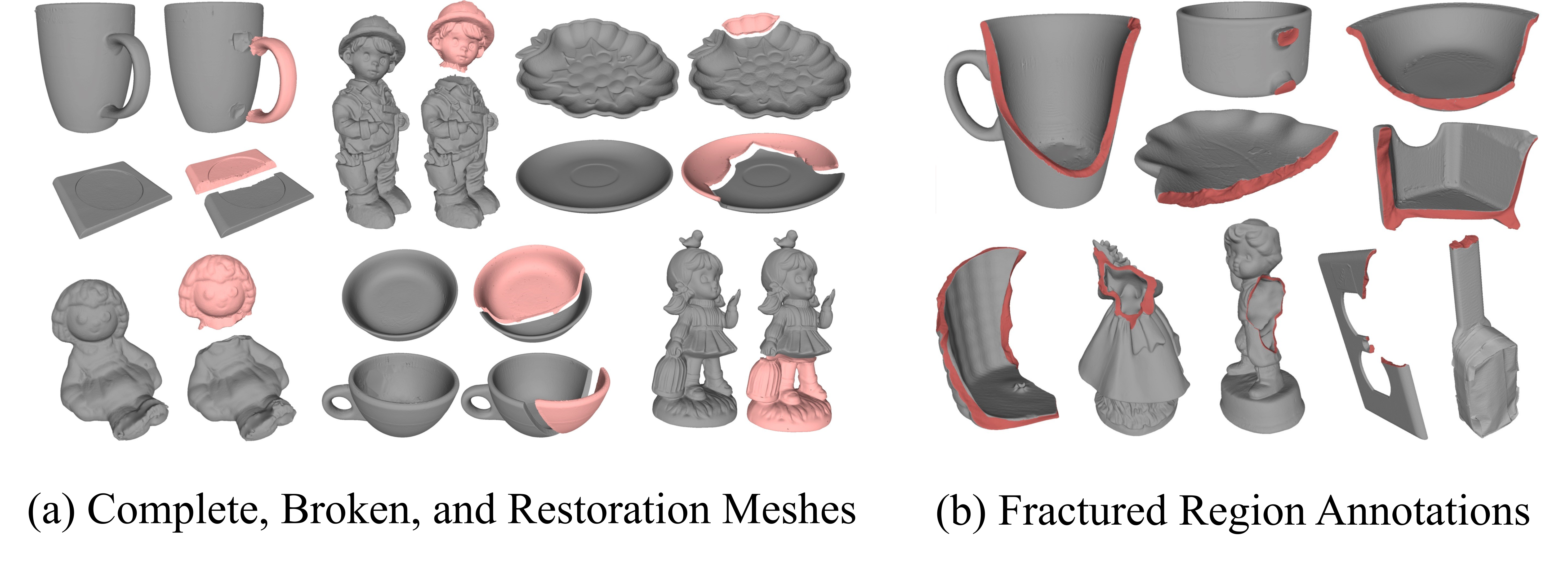}
    \caption{(a) For each triplet of meshes, we show the complete mesh on the left, the broken mesh aligned to the complete mesh on the right, and the restoration in pink detached from the broken. (b) Broken meshes (gray) with  labeled fracture surface (red).}
    \label{fig:demo}
\end{figure}

\textbf{Mesh Alignment.} Given the cleaned meshes for the broken object and its complete counterpart, we transform the broken mesh such that its intact regions are aligned with the corresponding regions of the complete mesh. We  perform an initial manual alignment of the broken mesh to the complete mesh, and refine the alignment using the iterative closest point (ICP)~\cite{besl1992method} algorithm. We conduct a post-alignment normalization of each mesh to ensure similarly scaled data for learning-driven repair. We perform normalization by scaling the broken and complete mesh such that the complete mesh is entirely contained within a unit cube. We provide access to non-normalized and normalized meshes as part of the dataset. Figure~\ref{fig:demo}(a) shows examples of the broken mesh aligned to the complete mesh.

\textbf{Ground Truth Restoration Estimation.} Our data collection involves scanning of broken objects that have either been acquired as pre-damaged, or have had damage inflicted through a destructive fracture process, such that only one part of the object remains intact. Such an occurrence is not uncommon, e.g., as shown in Figure~\ref{fig:fracturescan}, the fracture process causes the handle to further fragment into a number of small pieces that are infeasible to reassemble. However, shape repair approaches can benefit from geometric knowledge of ground truth parts needed to repair the object, in order to perform training and evaluation. We contribute proxy ground truth 3D meshes for the repair parts. We synthetically generate the repair part proxies using the approach of Lamb et al.~\cite{lamb2019automated}. Given aligned complete and broken meshes, the approach recovers restoration meshes that smoothly join the broken shape to repair the object, lack small-scale artifacts caused by Boolean subtraction, and lack grooves at the fracture boundary demonstrated by approximate subtraction techniques based on distance thresholding. Figure~\ref{fig:demo}(a) shows restoration meshes generated for cleaned and waterproofed broken and complete scans.

\textbf{Ground Truth Fracture Surface Annotation.} To assist with approaches that rely on accurate knowledge of the fracture surface for training or evaluation, we manually annotate triangles corresponding to the fracture surface. Figure~\ref{fig:demo}(b) shows example ground truth fracture annotations.

\section{Analysis of \textit{Fantastic Breaks}}
\label{sec:analysis}

\paragraph{Summary of Dataset.} We have annotated each object with a category level, its material, and the approach of damaging the object. At the time of publication we have acquired 214 and 241 physical damaged and complete objects, and 195 and 218 damaged and complete object scans. Among the 195 broken objects, 15 objects are naturally damaged, and the remaining have damage inflicted manually by dropping (26), striking using a mallet (126), striking using a hammer (25), and anchoring and snapping (3). 150 of the broken objects have a counterpart within the 241 complete objects. We have acquired pre-cleaned scanned meshes for 195 of the broken objects and 218 of the complete objects. 150 of the 214 broken scanned objects are paired with their respective 150 complete scanned objects. These 150 pairs are fully cleaned and have restorations extracted. Tables~\ref{tab:class} and \ref{tab:material} summarize physical, scanned, and paired scanned broken and complete objects by class and material respectively. This is a growing dataset, i.e., the physical and scanned sets are continuing to expand. We expect to have all 241 complete objects broken, and we continue to process complete and broken objects.

\setlength{\tabcolsep}{1pt}
\begin{table}[t!]
    \centering
    \caption{Object distribution by class (Phys.=Physical).}
    \footnotesize
    \begin{tabular}{@{}c|c|ccccccccc|c@{}}
    \toprule
    & Class & Mug & Plate & Statue & Bowl & Cup & Jar & Coaster & Box & Misc & Total\\
    \hline
   \multirow{2}{*}{\rotatebox{90}{Phys.}} & Broken & 46 & 43 & 39 & 23 & 12 & 13 & 6 & 6 & 26 & 214\\
    & Complete & 42 & 45 & 43 & 27 & 16 & 19 & 7 & 7 & 35 & 241\\
    \hline
   \multirow{3}{*}{\rotatebox{90}{Scan}} & Broken & 43 & 41 & 36 & 21 & 10 & 12 & 6 & 5 & 21 & 195\\
    & Complete & 35 & 44 & 41 & 24 & 14 & 15 & 7 & 7 & 31 & 218\\
    & Pairs & 30 & 35 & 30 & 17 & 8 & 6 & 6 & 3 & 15 & 150\\
\bottomrule
    \end{tabular}
    \label{tab:class}
\end{table}

\setlength{\tabcolsep}{2pt}
\begin{table}[t!]
    \centering
    \caption{Object distribution by material (Mat., Phys.=Physical).}
    \footnotesize
    \begin{tabular}{@{}c|c|cccccc|c@{}}
    \toprule
    & Mat. & Ceramic & Plastic & Glass & Plaster & Wood & Other & Total\\
    \hline
   \multirow{2}{*}{\rotatebox{90}{Phys.}} & Broken & 165 & 21 & 10 & 4 & 4 & 10 & 214\\
    & Complete & 170 & 30 & 10 & 3 & 8 & 20 & 241\\
    \hline
   \multirow{3}{*}{\rotatebox{90}{Scan}} & Broken & 155 & 17 & 6 & 4 & 4 & 9 & 195\\
    & Complete & 151 & 28 & 9 & 3 & 8 & 19 & 218\\
    & Pairs & 115 & 13 & 7 & 4 & 3 & 8 & 150\\
\bottomrule
    \end{tabular}
    \label{tab:material}
\end{table}

\paragraph{Analysis of Geometric Properties.}

\begin{figure*}[t!]
    \centering
    \includegraphics[width=\linewidth]{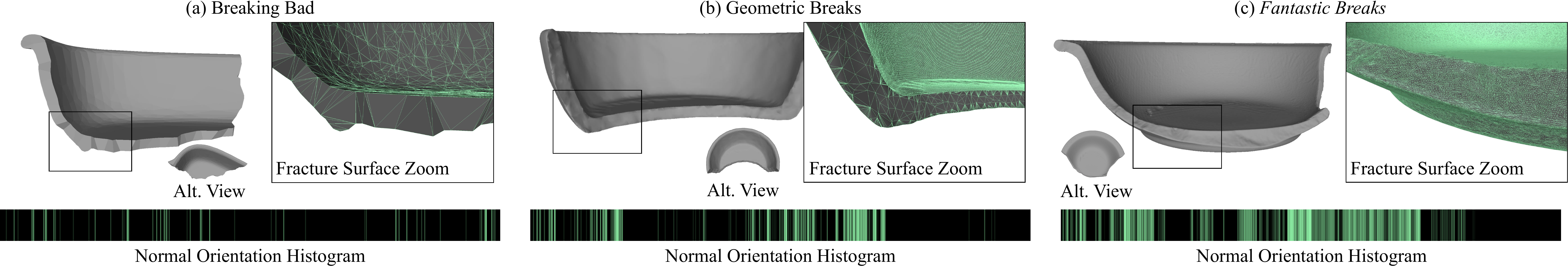}
    \caption{Broken shapes with inset showing geometric detail, alternate (alt.) top-down views, and normal orientation histograms (NOHs) for (a)~Breaking Bad, (b)~Geometric Breaks, and (c)~\textit{Fantastic Breaks}. NOHs are viewed as images where higher intensity lines represent higher bin counts. (a)~Breaking Bad shapes show sparse structure due to limitations on fracture resolution. (b)~While Geometric Breaks shapes have semi-dense structure, they demonstrate regularity attributed to the breaking primitive. (c)~\textit{Fantastic Breaks} shapes have dense surface structure and irregular break patterns characteristic of arbitrary real-world fracture.}
    \label{fig:norm_hist}
\end{figure*}

We quantitatively compare geometric properties of the \textit{Fantastic Breaks} dataset to the \texttt{everyday} subset of Breaking Bad~\cite{sellan2022breaking}, and the Geometric Breaks dataset provided by Lamb et al.~\cite{lamb2022deepjoin}, both of which contain synthetic fractures. The \texttt{everyday} subset of Breaking Bad contains 542 objects each fractured 100 times, for a total of 54,200 broken objects. Breaks are obtained using a physics driven fracturing technique in which a set of fracture modes are computed and used to simulate object fracture patterns that result from an impact to the object. The Geometric Breaks dataset contains 24,208 objects from ShapeNet~\cite{chang2015shapenet} and 1,042 objects from the GSO dataset. Objects are fractured by subtracting them with a randomized convex geometric primitive. For Breaking Bad we compute quantitative metrics over the 12,939 \texttt{everyday} objects that contain 2 broken parts. For Geometric Breaks, we compute quantitative metrics over all 25,250 objects. For \textit{Fantastic Breaks}, we compute quantitative metrics over the 150 paired broken objects.

We summarize the mean number of faces and vertices for broken meshes in each dataset on the left of Table~\ref{tab:quant}. Our broken meshes are, on average, at least ten times more dense than existing fractured object datasets. As observed by Sell\'{a}n et al.~\cite{sellan2022breaking}, an indicator of fractures generated using pre-computed fracture methods are a large number of convex fractured shapes. In the final three columns of Table~\ref{tab:quant} we provide the 25$^{\textrm{th}}$, 50$^{\textrm{th}}$, and 75$^{\textrm{th}}$ percentiles of broken shape convexity. We compute convexity as the ratio of the volume of the broken mesh to the volume of the mesh's convex hull, as used by Attene et al.~\cite{attene2008hierarchical}. Highly convex shapes show a convexity value near~1. Our dataset shows a consistently lower convexity than other datasets, with a 75$^{\textrm{th}}$ convexity percentile of 0.519, compared to 0.828 for Breaking Bad and 0.746 for Geometric Breaks.

\begin{table}[t!]
\centering
\small
\caption{Number of vertices, faces, and convexity percentiles for Breaking Bad, Geometric Breaks, and \textit{Fantastic Breaks} datasets. Max Vertices and Faces and min Convexity values are bolded.}
\begin{tabular}{@{}l|cc|ccc@{}}
\toprule
Dataset & \# Vertices & \# Faces & C 25th & C 50th & C 75th \\ \hline
Breaking Bad & 4,578 & 17,876 & 0.286 & 0.536 & 0.828 \\
Geometric & 52,719 & 105,436 & \textbf{0.229} & 0.480 & 0.746 \\
\textit{Fantastic} & \textbf{566,630} & \textbf{1,133,837} & 0.238 & \textbf{0.320} & \textbf{0.519} \\
\bottomrule
\end{tabular}
\label{tab:quant}
\end{table}

Though synthetically generated fractured objects may show coarse geometric variation, they struggle to generate fine detail at the fractured region. As shown in Figures~\ref{fig:norm_hist}(a) and \ref{fig:norm_hist}(b), synthetically generated fractured objects are characterized by piecewise fractured regions that lack high frequency surface variation, i.e. they do not capture the fine-scale variability of real fractures. Simulation of object fractures with fine-scale surface variability of the same resolution as real object fractures using a physics engine as done in Breaking Bad is intractable with current hardware. Fractures generated by subtracting geometric primitives demonstrate unnatural regularity as shown in Figure~\ref{fig:norm_hist}(b).

\begin{figure}[t!]
    \centering
    \includegraphics[width=\linewidth]{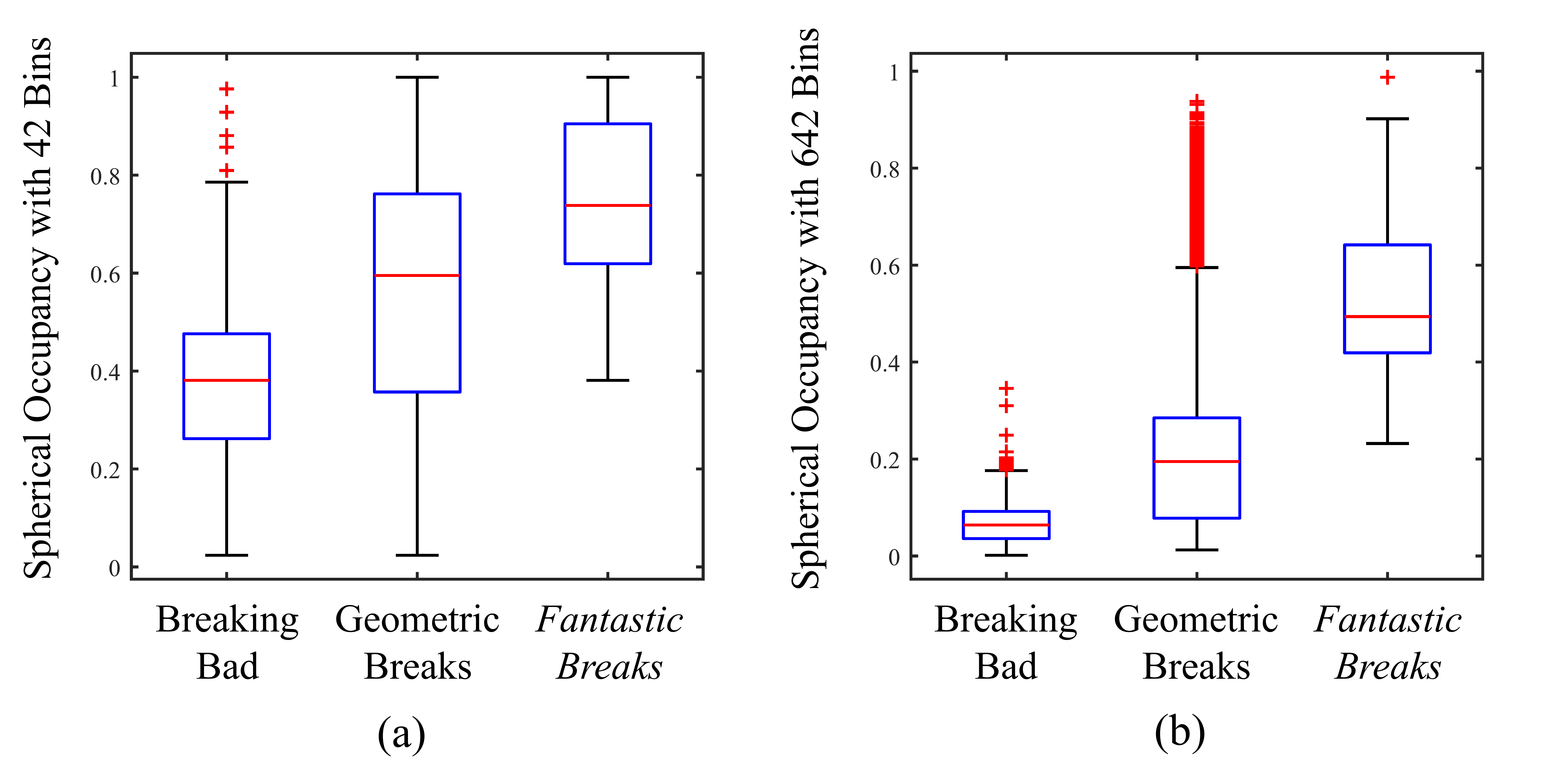}
    \caption{Box plots for distribution of spherical occupancy over objects from Breaking Bad, Geometric Breaks, and \textit{Fantastic Breaks} using (a) 42 bins and (b) 642 bins.}
    \label{fig:norm_box}
\end{figure}

To quantify coarse and fine-scale surface variability, we introduce the spherical occupancy metric, which measures the degree to which normals on the fracture surface occupy the space of all possible orientations when grouped into \textit{n} discrete bins. For a given broken object, we extract the fractured region of the broken mesh. We obtain \textit{n} evenly spaced points on the unit sphere, and bin the normal vectors of the fractured region into one of \textit{n} bins based on their closest point on the unit sphere to generate a normal orientation histogram (NOH) for the object. We show NOHs for three objects in Figure~\ref{fig:norm_hist}. To compute spherical occupancy, we count the number of occupied bins in the NOH and divide by \textit{n}. A fractured region with coarse geometry is expected to show low spherical occupancy, as its normals span a small portion of the space of all possible orientations. The choice for \textit{n} determines the scale of surface variability measured. A small value for \textit{n} measures coarse surface variability, i.e. highly concave or convex fractures will have a high spherical occupancy. A large value for \textit{n} measures fine-scale surface variability. If \textit{n} is large, only objects with fine-scale surface variability will show a high spherical occupancy, as the space of orientations is larger.

We show quantitative results for the spherical occupancy for two values of \textit{n} in Figure~\ref{fig:norm_box}(a) and Figure~\ref{fig:norm_box}(b). To measure coarse and fine surface variability, we set \textit{n} to 42 and 642 respectively. As shown in Figure~\ref{fig:norm_box}(a), most objects in \textit{Fantastic Breaks} have a high spherical occupancy when $\textit{n}=42$, indicating that objects in \textit{Fantastic Breaks} have a high degree of coarse surface variability. \textit{Fantastic Breaks} shows the highest mean spherical occupancy value of 0.747 when \textit{n} is 42, compared to 0.367 for Breaking Bad and 0.569 for Geometric Breaks. \textit{Fantastic Breaks} also contains mostly objects with high fine-scale surface variability, as shown by the lower spread in Figure~\ref{fig:norm_box}(b). \textit{Fantastic Breaks} shows the highest median spherical occupancy when \textit{n} is 642, indicating a large number of fractures with fine-scale surface variability, and includes the object with the highest spherical occupancy across all datasets. In contrast, the piecewise fractures of Breaking Bad and Geometric Breaks, demonstrate consistently lower occupancy, as shown in Figure~\ref{fig:norm_box}(b), indicating a lack of fine-scale surface variability. \textit{Fantastic Breaks} also has the highest mean spherical occupancy of 0.525 when $\textit{n}=642$, compared to 0.066 for Breaking Bad and 0.204 for Geometric Breaks.

\section{Experimental Evaluation}
\label{sec:exper}

As we provide complete, broken, and restoration shapes with ground truth fractured region annotations, our dataset may be used to train and evaluate approaches that perform automated shape repair. We test three prior shape repair approaches on our dataset: MendNet~\cite{lamb2022mendnet}, DeepMend~\cite{lamb2022deepmend}, and DeepJoin~\cite{lamb2022deepjoin}. These approaches generate repair parts assuming that the missing part has been lost or destroyed during the damage process. MendNet, DeepMend, and DeepJoin represent shapes by learning a function to reconstruct shapes as implicit surfaces, and require watertight 3D meshes as input. DeepJoin also requires meshes to have surface normals, which \textit{Fantastic Breaks} provides.

For shape repair methods, we pre-train a given network on the Geometric Breaks dataset or Breaking Bad dataset. After training for 2,000 epochs on synthetically fractured objects, we train for an additional 1,000 epochs on objects from the \textit{Fantastic Breaks} dataset. To generate training data, for each broken, complete, and restoration mesh tuple in each dataset, we sample points on the surface of each mesh and compute the signed distance function (SDF), occupancy, and normal field value for each point. For DeepMend and DeepJoin we compute a break surface that acts as a proxy for the fracture by fitting a thin-plate spline to the fracture vertices, as described by Lamb et al.~\cite{lamb2022deepjoin}. We train with 5,068 objects from the mugs, jars, bottles, and GSO classes from Geometric Breaks, and hold out the remaining 45 for evaluation. For Breaking Bad, we train with \texttt{everyday} objects that have 2 parts. Breaking Bad objects are not waterproof, so we perform waterproofing before training. We train with 2,575 objects from the \texttt{everyday} subset of Breaking Bad that are successfully waterproofed, and hold out the remaining 45 for evaluation. \textit{Fantastic Breaks} has 105 train and 45 test objects.

\begin{figure}[t!]
    \centering
    \includegraphics[width=\linewidth]{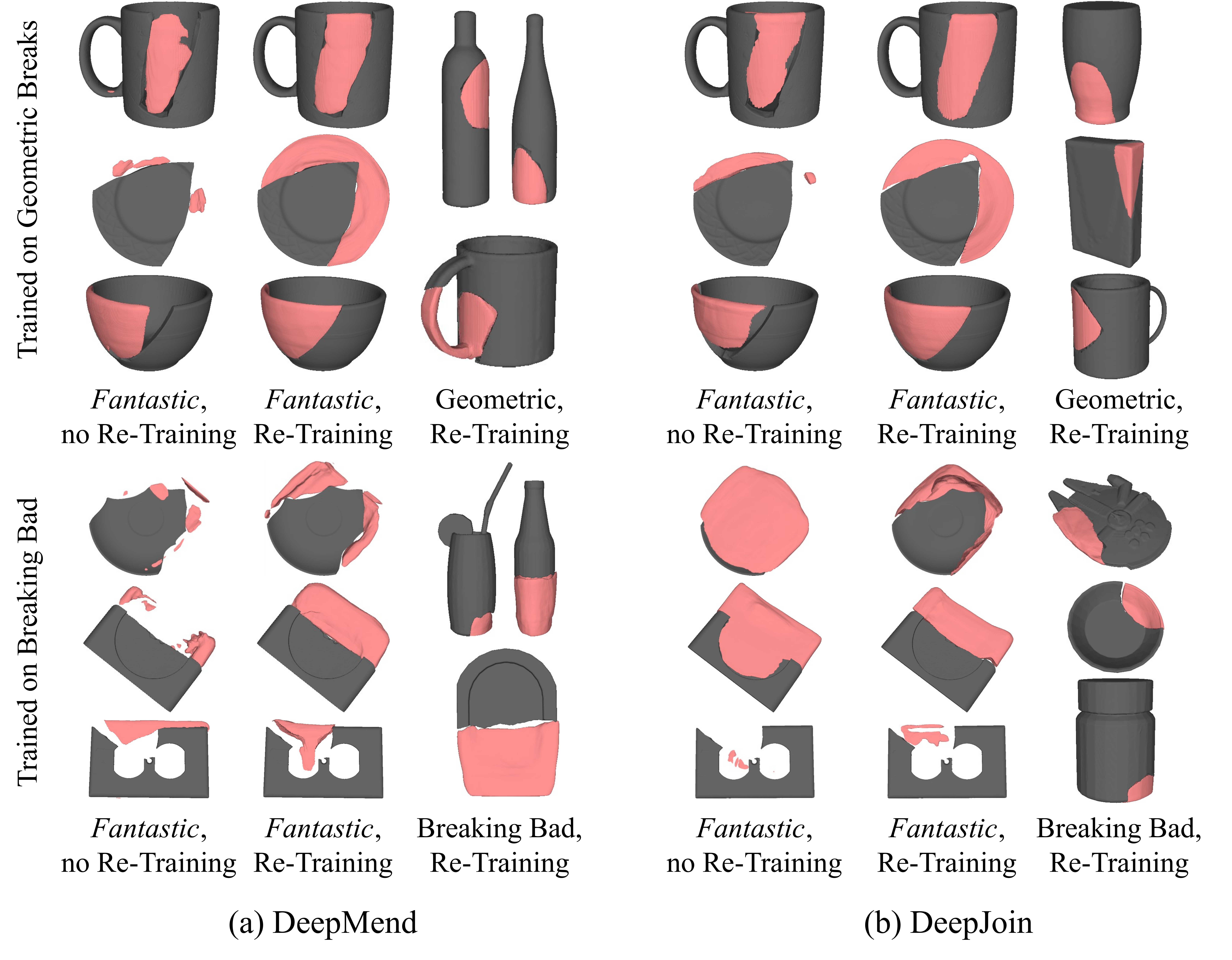}
    \caption{Predicted repair shapes in pink trained with (top) Geometric Breaks and (bottom) Breaking Bad using (a) DeepMend and (b) DeepJoin. \textit{Fantastic Breaks} objects are shown before and after re-training, which improves results for Geometric Breaks.}
    \label{fig:repair}
\end{figure}

Figure~\ref{fig:repair} shows repairs for broken objects before and after re-training on \textit{Fantastic Breaks}. Before re-training on real objects, DeepMend and DeepJoin may struggle to generate repairs that fully restore real objects, e.g. the plates and mugs on the left of Figures~\ref{fig:repair}(a)~and~(b). Re-training on \textit{Fantastic Breaks} improves repairs for some objects, as shown in the middle of Figure~\ref{fig:repair}, producing more holistic repairs. As shown in the right columns of Figure~\ref{fig:repair}, re-training still allows generation of repairs for synthetic fractures. 

To measure the accuracy of predicted repairs we use the chamfer distance (CD), given by Park et al.~\cite{park2019deepsdf}, and normal consistency (NC), given by Mescheder et al.~\cite{mescheder2019occupancy}. As MendNet, DeepMend, and DeepJoin perform optimization during inference to obtain repairs, inference proceeds non-deterministically. For each approach, we pre-train Geometric Breaks or Breaking Bad, and perform re-training on \textit{Fantastic Breaks}. For evaluation before re-training, we report quantitative metrics over 21 inference runs without re-training, totaling 21 trials. For evaluation after re-training, we report quantitative metrics over 3 re-trained models and 7 inference runs with re-training, totalling 21 trials.

Table~\ref{tab:repair} shows success rate, CD, and NC for predicted repairs using MendNet, DeepMend, and DeepJoin. As shown in the top of Table~\ref{tab:repair}, re-training on the \textit{Fantastic Breaks} dataset decreases CD and increases NC of repair parts on Geometric Breaks for DeepMend and Deepjoin. As acknowledged by the authors, MendNet struggles to restore real objects when trained on synthetic fractures. Thus re-training on 105 real samples may not benefit learning. Figures~\ref{fig:repair}(a)~and~(b) show an improvement for the mug, plate, and cup from Geometric Breaks after re-training. Table~\ref{tab:repair} shows a decrease in CD from 0.059 to 0.051 and from 0.030 to 0.022 for DeepMend and DeepJoin respectively after re-training. Breaking Bad contains irregular objects, such as the tombstone and the model ship in Figures~\ref{fig:repair}(a)~and~(b), which may have more than half of the object fractured off, making it a challenging dataset for shape repair. Though re-training may improve performance on some real objects, e.g. the coaster in Figure~\ref{fig:repair}, overall re-training on \textit{Fantastic Breaks} does not improve performance, as shown on the bottom of Table~\ref{tab:repair}. However, evaluation on our dataset demonstrates that, though all three repair approaches generate feasible restoration shapes for repairing purely synthetic fractures, models trained on Breaking Bad generalize less well to real fractured objects than models trained on Geometric Breaks, with a CD of 0.080 and 0.110 with MendNet, 0.059 and 0.065 with DeepMend, and of 0.030 and 0.061 with DeepJoin for Geometric Breaks and Breaking Bad respectively, as shown in the left supercolumn of Table~\ref{tab:repair}.

\setlength{\tabcolsep}{4pt}
\begin{table}[t!]
\caption{Success rate (SR), chamfer distance (CD), and normal consistency (NC), with training on Geometric Breaks (GB, top) and on Breaking Bad (BB, bottom), before (left) and after (right) re-training on \textit{Fantastic Breaks} (FB).}
    \centering
    \small
    \begin{tabular}{@{}cc|ccc|ccc@{}}
    \toprule
    & & \multicolumn{3}{c|}{Train with GB Only} & \multicolumn{3}{c}{Train with GB + FB}\\
    & Test & SR\% & CD & NC & SR\% & CD & NC\\
    \hline
    \multirow{2}{*}{\rotatebox{90}{$\begin{matrix}\textrm{Mend}\\\textrm{Net}\end{matrix}$}}& 
      GB & \textbf{99.89\%}  & \textbf{0.132} & \textbf{0.198} & \textbf{99.89\%}  & 0.205          & 0.185          \\
& FB & \textbf{99.68\%}  & \textbf{0.080} & 0.276          & 99.58\%           & 0.094          & \textbf{0.277} \\
    \hline
    \multirow{2}{*}{\rotatebox{90}{$\begin{matrix}\textrm{Deep}\\\textrm{Mend}\end{matrix}$}}& 
      GB & 95.45\%           & \textbf{0.063} & \textbf{0.607} & \textbf{93.65\%}  & 0.085          & 0.548          \\
& FB & 98.20\%           & 0.059          & 0.323          & \textbf{99.05\%}  & \textbf{0.051} & \textbf{0.441} \\
    \hline
    \multirow{2}{*}{\rotatebox{90}{$\begin{matrix}\textrm{Deep}\\\textrm{Join}\end{matrix}$}}& 
      GB & 92.80\%           & \textbf{0.064} & \textbf{0.658} & \textbf{95.98\%}  & 0.102          & 0.526          \\
& FB & \textbf{100.00\%} & 0.030          & 0.426          & 99.89\%           & \textbf{0.022} & \textbf{0.504} \\
    \midrule
    \midrule
    & & \multicolumn{3}{c|}{Train with BB Only} & \multicolumn{3}{c}{Train with BB + FB}\\
    & Test & SR\% & CD & NC & SR\% & CD & NC\\
    \hline
    \multirow{2}{*}{\rotatebox{90}{$\begin{matrix}\textrm{Mend}\\\textrm{Net}\end{matrix}$}} & 
      BB & 93.23\%           & \textbf{0.122} & \textbf{0.344} & \textbf{100.00\%} & 0.210          & 0.307          \\
& FB & 95.13\%           & \textbf{0.110} & 0.184          & \textbf{100.00\%} & 0.116          & \textbf{0.302} \\
    \hline
    \multirow{2}{*}{\rotatebox{90}{$\begin{matrix}\textrm{Deep}\\\textrm{Mend}\end{matrix}$}} & 
      BB & \textbf{96.19\%}  & \textbf{0.078} & \textbf{0.637} & 92.70\%           & 0.176          & 0.351          \\
& FB & 99.68\%           & \textbf{0.065} & \textbf{0.213} & \textbf{100.00\%} & 0.092          & 0.190          \\
    \hline
    \multirow{2}{*}{\rotatebox{90}{$\begin{matrix}\textrm{Deep}\\\textrm{Join}\end{matrix}$}}& 
      BB & 98.94\%           & \textbf{0.042} & \textbf{0.688} & \textbf{100.00\%} & 0.177          & 0.285          \\
& FB & \textbf{100.00\%} & \textbf{0.061} & \textbf{0.259} & \textbf{100.00\%} & 0.076          & 0.203         \\
    \bottomrule
    \end{tabular}
    \label{tab:repair}
\end{table}
\section{Discussion} 
\label{sec:discussion}

We present \textit{Fantastic Breaks}, a novel dataset that contains full 360$^\circ$ 3D scans of broken objects geometrically aligned with 3D scans of their complete counterparts, with manual annotations of classes, materials, and fracture surfaces, and synthetic proxies for 3D meshes representing repair parts. The dataset continues to grow in number of physical objects and scans. The dataset is one of the first of its kind, enabling learning of the characteristics of fractured objects. \textit{Fantastic Breaks} provides data-driven insight into fracture, overcoming the deficits of geometric approaches that make prior assumptions about the damage process that are not widely applicable, as well as the concerns of datasets based on physics simulations that are limited by current hardware in modeling real-world geometry.

An obvious limitation of our endeavor is that employing destructive processes to damage objects for the purpose of real-world data acquisition is unsustainable at a large scale. We advocate that our dataset be leveraged to learn patterns of break and internal geometric structure that are common across objects of similar materials, such as ceramics or plastics, and classes, such as mugs or cups, and to use generative approaches to conduct data-driven synthesis of breaks and internal structure given 3D models of complete objects. For instance, an interesting observation of our collection is that for reasons of cost, sustainable production, and functionality, nearly all our objects have shell rather than solid structures, whereas scans of whole objects such as bottles or statues cannot capture the internal shell structure. By exposing the internal structure, the dataset provides the opportunity to learn how to hollow out 3D models of objects, an opportunity absent from prior 3D scan datasets.

\textit{Fantastic Breaks} currently contains tabletop objects aquired using desktop scanners. Larger objects may necessitate more elaborate setups, e.g., room-scale imaging systems. Future data collection can investigate the minimal number of viewpoints needed to acquire geometrically relevant understanding of internal object structure. For instance, to capture the fracture pattern of a chair leg, it may be sufficient to image the broken region from 1-2 viewpoints using a depth camera, and obtain and deform a 3D proxy from a public repository to the imaged viewpoints.

We have provided evaluations of existing approaches on automatic reconstruction of new repair parts using learning-driven approaches. The dataset is widely applicable to a range of other tasks, e.g., the broken and restoration meshes can be used to perform shape assembly cognizant of precision joins for real-world fracture boundaries. We evaluate a shape repair approach that does not require prior knowledge of the fractured region (MendNet). However, via our manual fracture surface annotations, we also plan to release the incomplete meshes devoid of the fracture surface. These incomplete meshes will benefit research in partial shape completion~\cite{fei2022comprehensive}, which up until now has largely focused on synthetically generated partial shapes or on depth scans. Datasets of real-world 3D scans have contributed to significant advancements in robotic manipulation~\cite{fang2020graspnet,chao2021dexycb}. \textit{Fantastic Breaks} provides impacts in robotics research by enabling robot-driven repair, object grasp while being cognizant of fractured regions to minimize further damage, and damaged object handling for safe human-robot handover. 

\paragraph{Acknowledgments}
This work is supported by the National Science Foundation (NSF) under grant IIS-2023998. We would like to thank Joseph Morrison, Nikolai Melnikov, Odin Kohler, Patrick O'Mahony, Holly Rossmann, Rosalina Delwiche, Elek Ye, and Noah Wiederhold for their assistance with object scanning.

%%%%%%%%% REFERENCES
{\small
\bibliographystyle{ieee_fullname}
\bibliography{references}
}

\end{document}